%% file: aaai2026.tex
\documentclass[letterpaper]{article} 

\ifdefined\aaaianonymous
    \usepackage[submission]{aaai2026}  
\else
    \usepackage{aaai2026}              
\fi

\usepackage{times}  
\usepackage{helvet}  
\usepackage{courier}  
\usepackage[hyphens]{url}  
\usepackage{graphicx} 
\usepackage{natbib}  
\usepackage{caption} 
\usepackage{amsmath}
\usepackage{amsfonts} 
\usepackage{graphicx}  
\usepackage{booktabs}  

\usepackage{algorithm}
\usepackage{algorithmic}
\usepackage{times}
\usepackage{helvet}
\usepackage{courier}
\usepackage{xcolor}
\usepackage{newfloat}
\usepackage{listings}
\DeclareCaptionStyle{ruled}{labelfont=normalfont,labelsep=colon,strut=off} 
\floatstyle{ruled}
\newfloat{listing}{tb}{lst}{}
\floatname{listing}{Listing}

\ifdefined\aaaianonymous
    \title{Dream3DAvatar: Text-Controlled 3D Avatar Reconstruction from a Single Image}
\else
    \title{Dream3DAvatar: Text-Controlled 3D Avatar Reconstruction from a Single Image}
\fi

\author {
    Gaofeng Liu\textsuperscript{\rm 1}\footnotemark[1]~,
    Hengsen Li\textsuperscript{\rm 1}\footnotemark[1]~,
    Ruoyu Gao\textsuperscript{\rm 1},
    Xuetong Li\textsuperscript{\rm 1},
    Zhiyuan Ma\textsuperscript{\rm 2}\footnotemark[2]~,
    Tao Fang\textsuperscript{\rm 1}\footnotemark[2]~ 
}

\affiliations {
    \textsuperscript{\rm 1}Department of Automation, Shanghai Jiao Tong University, Shanghai 201100, China.\\
    \textsuperscript{\rm 2}Department of Electronic Engineering, Tsinghua University, Beijing 100084, China.\\
}

\usepackage{bibentry}

\begin{document}


\maketitle

\input{tex/0_abstract}

\input{tex/1_introduction}
\input{tex/2_related_work}
\input{tex/3_method}

\input{tex/4_experiments}

\input{tex/5_conclusion}

\appendix

\bibliography{aaai2026}

\clearpage
\appendix
\end{document}

%% file: tex/0_abstract.tex
\begin{abstract}
With the rapid advancement of 3D representation techniques and generative models, substantial progress has been made in reconstructing full-body 3D avatars from a single image. However, this task remains fundamentally ill-posedness due to the limited information available from monocular input, making it difficult to control the geometry and texture of occluded regions during generation.
To address these challenges, we redesign the reconstruction pipeline and propose Dream3DAvatar, an efficient and text-controllable two-stage framework for 3D avatar generation.
In the first stage, we develop a lightweight, adapter-enhanced multi-view generation model. Specifically, we introduce the Pose-Adapter to inject SMPL-X renderings and skeletal information into SDXL, enforcing geometric and pose consistency across views. To preserve facial identity, we incorporate ID-Adapter-G, which injects high-resolution facial features into the generation process. Additionally, we leverage BLIP2 to generate high-quality textual descriptions of the multi-view images, enhancing text-driven controllability in occluded regions.
In the second stage, we design a feedforward Transformer model equipped with a multi-view feature fusion module to reconstruct high-fidelity 3D Gaussian Splat representations (3DGS) from the generated images. Furthermore, we introduce ID-Adapter-R, which utilizes a gating mechanism to effectively fuse facial features into the reconstruction process, improving high-frequency detail recovery.
Extensive experiments demonstrate that our method can generate realistic, animation-ready 3D avatars without any post-processing and consistently outperforms existing baselines across multiple evaluation metrics.
\end{abstract}

%% file: tex/1_introduction.tex
\section{Introduction}
\begin{figure}[t]
\centering
\includegraphics[width=1.0\columnwidth]{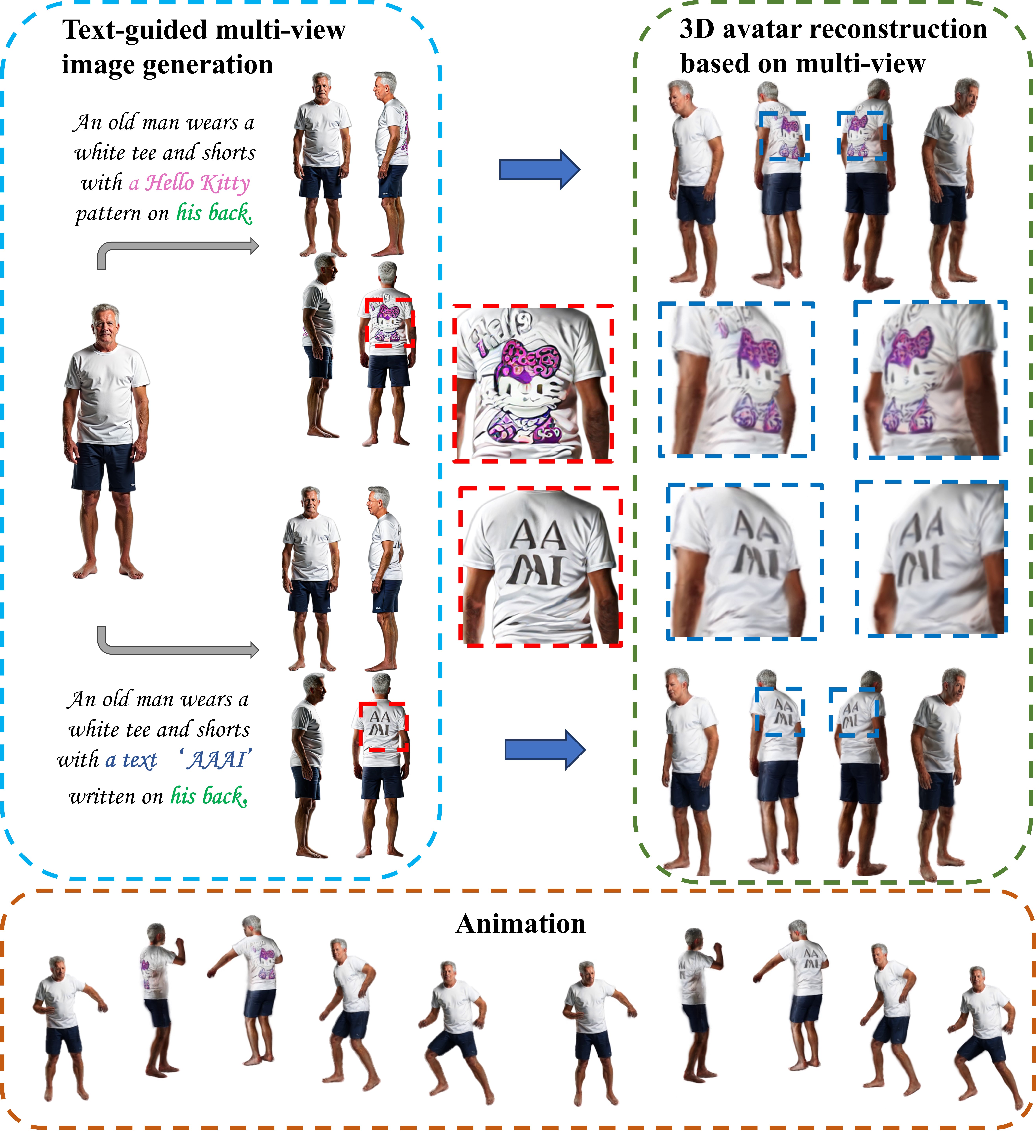}
\caption{We introduces a two stage animatable 3D human reconstruction method with text control from single image.}
\label{img:first_show_img}
\vspace{-10pt}
\end{figure}
3D digital humans have been widely applied in various fields such as gaming, animation, and virtual reality. However, creating realistic 3D avatars typically requires highly specialized modeling expertise and substantial manual labor. To alleviate this burden, recent research has increasingly focused on reconstructing 3D humans from a single image~\cite{li2025pshuman,zhuang2025idol,qiu2025lhm,han2023high}. Despite these approaches reduce the modeling cost, they struggle to provide text-driven control over the textures or geometry of occluded regions in the input image. This lack of controllability introduces a fundamental ill-posedness in the reconstruction process: the results heavily rely on prior knowledge learned by the model and represent only one of many possible interpretations. Given the inherent information loss in monocular images and the complexity of human pose, geometry, and texture, generating multi-view-consistent and photorealistic 3D avatars from a single image using a text-driven approach remains a highly challenging task.~\cite{saito2019pifu,xiu2023econ,saito2020pifuhd,zhang2024sifu,zheng2021pamir}. Early parametric models~\cite{alldieck2018video,loper2023smpl} leverage human priors to reconstruct 3D bodies but produce only coarse surface textures. Feedforward-based methods~\cite{zhuang2025idol,qiu2025lhm,zhang20243gen} allow for fast 3D reconstruction from a single image but lack controllability and diversity. Although recent advances such as diffusion-based models and iterative optimization~\cite{albahar2023single,li2025pshuman,xiu2022icon,choi2025svad} significantly improve reconstruction quality by exploiting multi-view information, they still suffer from low efficiency, limiting their practicality for real-time or high-throughput applications.

To tackle these challenges, we propose Dream3DAvatar, a text-driven, lightweight two-stage framework for reconstructing 3D virtual humans from a single image. Our method primarily addresses two key issues:~\textbf{1) alleviating the ill-posedness caused by monocular input; and 2) achieving lightweight and efficient 3D human reconstruction based on multi-view images.}

We incorporate three lightweight Adapter modules~\cite{alayrac2022flamingo} into the existing generative pipeline~\cite{podell2023sdxl,he2022masked}, enabling efficient fine-tuning while preserving the prior knowledge of the pretrained model. In the first stage, inspired by MV-Adapter~\cite{huang2024mv}, we leverage SDXL~\cite{podell2023sdxl} to controllably generate multi-view head images from a single input photo. To impose geometric consistency on body shape and pose, we introduce a geometry-aware module, Pose-Adapter, which injects SMPL-X and skeletal information as additional conditions into SDXL. Simultaneously, we employ ID-Adapter-G to process high-resolution facial features and ensure identity consistency across views. These conditions are fused within SDXL through multiple parallel attention mechanisms, facilitating multi-view image generation with consistent attributes. Additionally, we utilize BLIP2~\cite{li2023blip} to generate textual descriptions for the multi-view images, enabling text-conditioned training and enhancing controllability in occluded regions.

In the second stage, we employ a feedforward Transformer model equipped with a multi-view feature fusion module~\cite{zhuang2025idol} to reconstruct high-fidelity 3D virtual humans from the generated multi-view images. To better preserve identity features, we embed ID-Adapter-R into the Transformer architecture and introduce a gating mechanism to enhance the integration of facial details into the full-body reconstruction. With the incorporation of these lightweight modules, our method requires fine-tuning only an additional 80M parameters—approximately one-fifteenth of the total parameters in the Transformer model—greatly reducing the training cost. Furthermore, benefiting from a unified 3D human representation, our framework can effortlessly generate animated characters in diverse poses.

Dream3DAvatar combines the diversity of diffusion-based generation with the deterministic efficiency of feedforward Transformers, enabling fast and controllable 3D human reconstruction. We conduct both quantitative and qualitative comparisons with other methods in single-image to multi-view generation and multi-view to 3D reconstruction tasks. Results demonstrate that Dream3DAvatar consistently delivers superior performance across various challenging scenarios. In summary, our main contributions are as follows:
\begin{itemize}
    \item We propose \textbf{Dream3DAvatar}, a two-stage framework for efficient and text-controllable 3D human reconstruction from a single image, addressing key challenges of monocular ambiguity and limited controllability.
    \item We design an SDXL-based module that generates multi-view images with consistent texture and geometry from a single image, guided by text.
    \item We introduce a feed-forward transformer model that fuses facial features and multi-view features for high-quality 3D avatars reconstruction.
\end{itemize}






%% file: tex/2_related_work.tex
\begin{figure*}[t]
\centering
\includegraphics[width=1.0\textwidth]{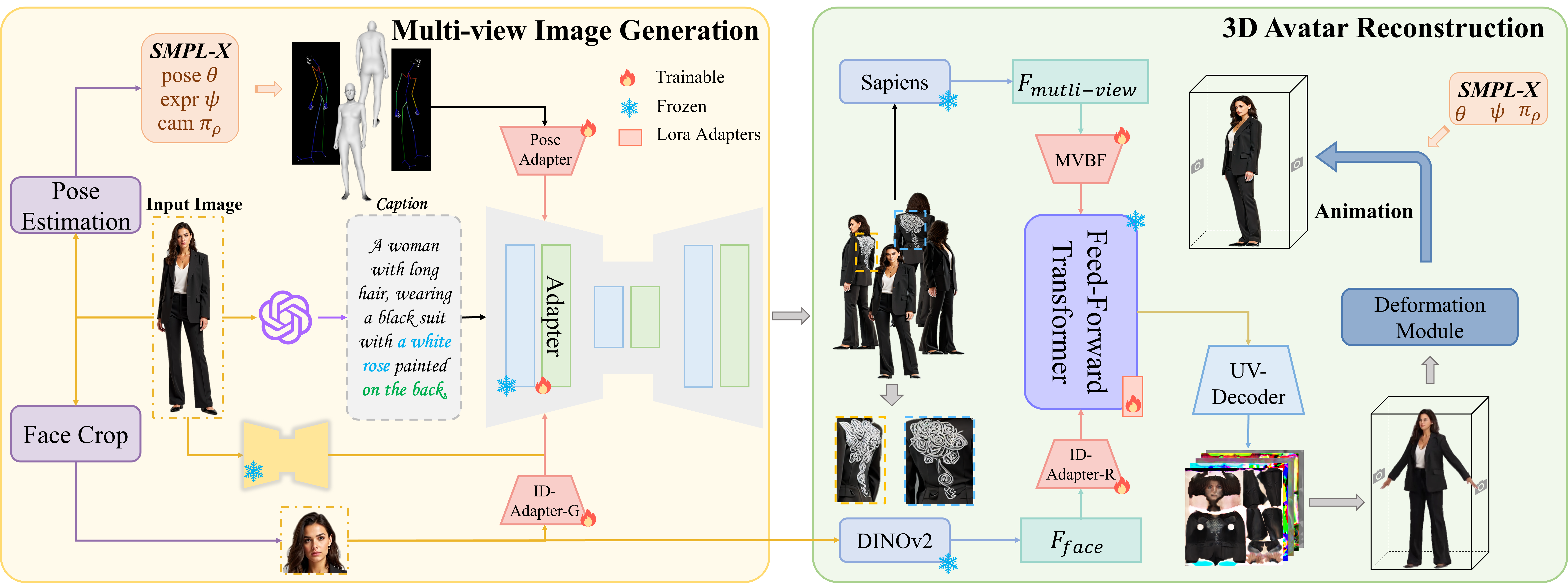} 
\caption{The overall framework of \textbf{Dream3DAvatar}.~\textbf{(Left)} A diffusion model that generates multi-view images from a single input image, incorporating SMPL-X geometry, skeleton and high-resolution facial images, and using Pose-Adapter and ID-Adapter-G modules to achieve pose consistency and identity preservation.~\textbf{(Right)} A feed-forward Transformer model that reconstructs high-fidelity 3D avatars from the generated multi-view images, significantly enhancing detail recovery and identity consistency through multi-view feature fusion and the ID-Adapter-R module.}
\label{fig:fig2}
\vspace{-14pt}
\end{figure*}
\section{Related Work}
\subsection{Stable Diffusion for Multi-View Image Generation}

Recent advances have adapted Stable Diffusion for novel-view synthesis. Early works~\cite{shi2023zero123++,shi2023mvdream} introduced camera-conditioned diffusion to generate plausible views from a single image. Later efforts~\cite{liu2023syncdreamer,shi2023toss,tang2024mvdiffusion++,li2024era3d,wu2024unique3d,long2024wonder3d} improved cross-view consistency and resolution. MV-Adapter~\cite{huang2024mv} proposed a lightweight adapter for efficient fine-tuning, balancing quality and speed. However, these generic models struggle to capture articulated human body structures.

To address this, human-specific models have been developed. Bhunia et al.~\cite{bhunia2023person} used pose conditioning to handle large articulations. Others~\cite{shao2024human4dit,kant2025pippo} adopted transformer-based diffusion for multi-view synthesis. Liu et al.~\cite{liu2024human} extended diffusion to videos. MagicMan~\cite{he2025magicman} incorporated SMPL and normal maps to produce coherent full-body views, but lacked facial detail. PSHuman~\cite{li2025pshuman} introduced separate body and face branches for improved facial fidelity, yet at high computational cost and without explicit control over occlusions.


\subsection{Single-Image 3D Human Reconstruction}

Reconstructing 3D humans from a single image is inherently ill-posed due to occlusions and missing views. Implicit methods~\cite{saito2019pifu,saito2020pifuhd} learn pixel-aligned occupancy or SDFs for detail recovery but require dense supervision and struggle with complex poses. Tri-plane representations~\cite{wang2023rodin,zhang2024rodinhd} improve speed and training efficiency, but suffer from weak priors and oversmoothing in unseen regions.

SMPL-based approaches~\cite{loper2023smpl,pavlakos2019expressive,xiu2022icon,xiu2023econ} introduce pose and topology constraints, enhancing robustness but inheriting fitting artifacts. Feed-forward methods~\cite{qiu2025lhm,zhuang2025idol,zhang20243gen} directly predict animatable 3D humans from a single view, offering real-time inference but lacking view-aware reasoning.

Recent methods~\cite{li2025pshuman,he2025stdgen,pan2024humansplat,ho2024sith,weng2024template} hallucinate pseudo-views to reduce ambiguity, though often with limited control and view inconsistency. AniGS~\cite{qiu2025anigs} generates multi-view sequences via video diffusion and fuses them into 4D Gaussian splats, achieving coherence but with efficiency trade-offs.

Our two-stage framework first synthesizes consistent multi-view images, then reconstructs a controllable 3D Gaussian Splatting representation, achieving both high fidelity and editability from a single image.

%% file: tex/3_method.tex
\section{Method}
\subsection{Overview}
We propose Dream3DAvatar, a two-stage framework for reconstructing high-fidelity 3D avatars from a single image (Fig. \ref{fig:fig2}). The pipeline comprises: (1) a multi-view image generation stage leveraging SDXL~\cite{podell2023sdxl} with Pose-Adapter and ID-Adapter-G modules to produce consistent multi-view images from single input, where Pose-Adapter enforces geometric consistency using SMPL-X and skeletal priors while ID-Adapter-G preserves identity through high-resolution facial features, guided by text prompts for occluded regions; and (2) a 3D avatar reconstruction stage employing a feedforward Transformer~\cite{zhuang2025idol} with multi-view feature fusion to generate 3D Gaussian splats (3DGS), enhanced by ID-Adapter-R's gating mechanism for facial detail recovery. 

Section \ref{sec:multi-view_image_generation} details the multi-view generation, Section \ref{sec:Multi-view_Transformer_for_3D_Recovery} describes the reconstruction model, and Section \ref{sec:Training_Strategy} explains the training methodology.


\subsection{Multi-view Image Generation}
\label{sec:multi-view_image_generation}
\subsubsection{Geometric and Semantic Conditioning}
\label{sec:Geometric_and_Semantic_Conditioning}
As shown on the left side of Figure~\ref{fig:fig2}, we augment the pre-trained SDXL model with several Adapters to generate high-fidelity, consistent multi-view images from a single person image. Specifically, we train a diffusion model capable of encoding SMPL-X images, skeletons, high-resolution facial features, and text, achieving consistency constraints from multiple angles such as geometry, pose, and texture for multi-view images.

\textbf{SMPL-X Guider.} Training a diffusion model to generate multi-view images from a single person image while learning pose information is a significant challenge. To address this, we introduce the Pose-Adapter, which encodes multi-view SMPL-X images and injects them into the SDXL model, enforcing consistency on the generated multi-view avatars from the perspectives of geometry and pose. These multi-view SMPL-X images are estimated from a single input image using a pre-trained pose estimation model, then rendered according to specified camera parameters.

\textbf{Skeleton Guider.} While SMPL-X provides good geometric consistency across multi-view images, it fails to maintain pose consistency for joints such as fingers. Therefore, we utilize the Pose-Adapter to inject body skeleton information into the model. The body skeleton is calculated using the SMPL-X parameters.

\textbf{Face Guider.} The face region occupies only a small portion of the image, yet its reconstruction quality is crucial for preserving the identity and overall reconstruction accuracy. Thus, we extract high-resolution facial features from the input single-view image and inject them into the diffusion model using the ID-Adapter-G, thereby enhancing the quality of facial region reconstruction.

\textbf{Text Guider.} During training, we employ an advanced Vision-Language Model (VLM) to extract descriptive text from the person image as an additional conditioning input. While the aforementioned conditions allow SDXL to generate reasonably coherent multi-view avatars, the texture of occluded regions in a single-view image might not meet user expectations due to the partial information it contains. As shown in Figure~\ref{fig:Tetx_Control}, by leveraging text conditions, we gain control over the texture details in the unseen region of generated multi-view images, ensuring that the textures align with the desired characteristics.


\begin{figure}[t]
\centering
\includegraphics[width=1.0\columnwidth]{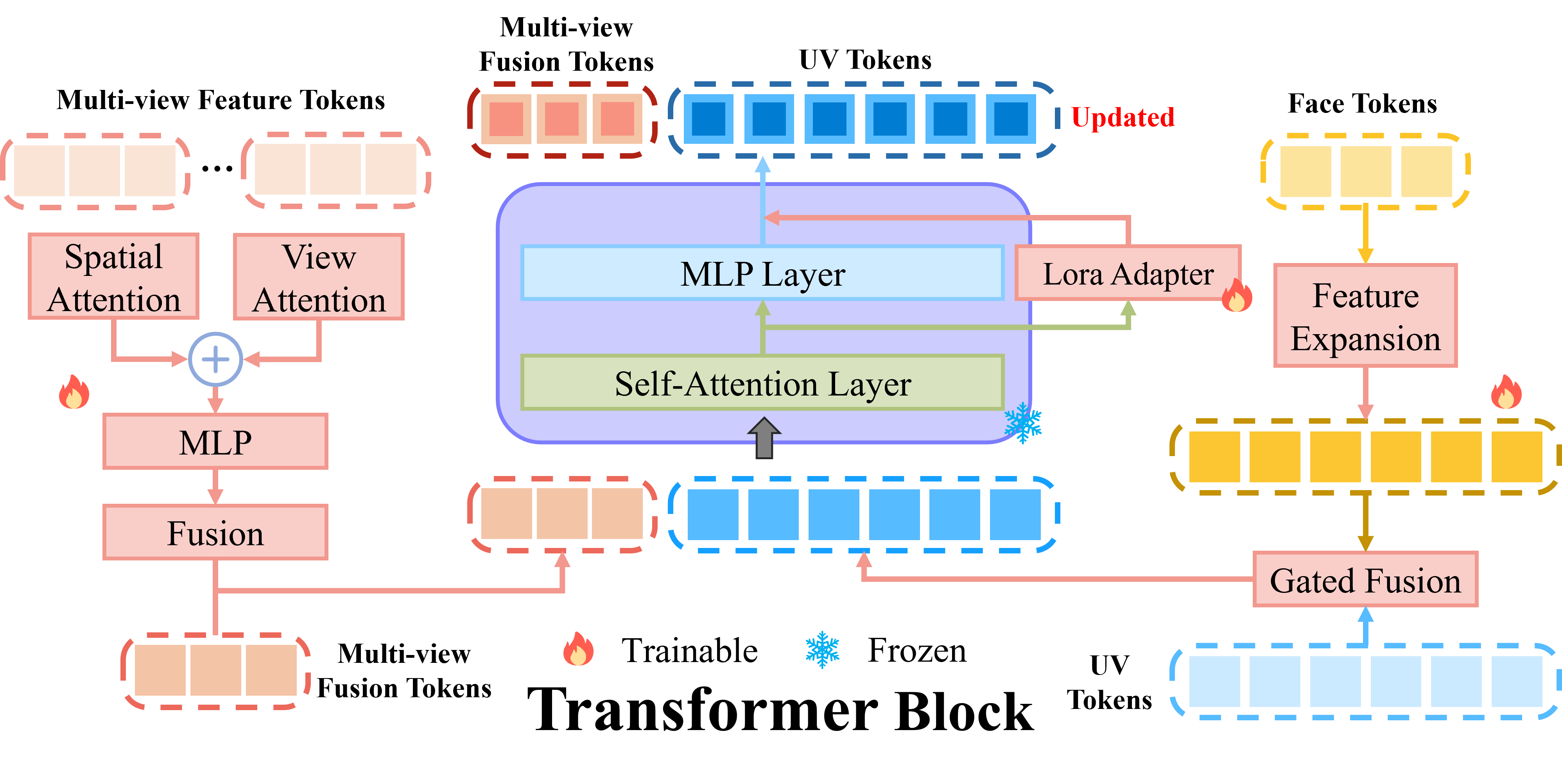}
\caption{The feedforward transformer block.}
\label{img:multi_view_fusion}
\vspace{-10pt}
\end{figure}
\subsubsection{Adapter-Enhanced Diffusion model}
\label{sec:Adapter-Enhanced_Diffusion_model}

We design the Pose-Adapter and ID-Adapter-G to inject human pose information, and facial information into the pre-trained SDXL model to generate high-fidelity, consistent multi-view images while retaining the structure and prior knowledge of the pre-trained SDXL model.

\textbf{Pose-Adapter.} The Pose-Adapter extracts multi-scale pose features from multi-view SMPLX images and skeleton images, and injects them into the encoder of the pre-trained UNet to add pose constraints. Considering that we use two pose-guided conditions, which contain different characteristics of the human body, we are inspired by~\cite{zhu2024champ} and introduce self-attention for each pose condition. This allows the model to capture the semantic information of different pose conditions. In each layer of the pose adapter, the two pose features are aggregated through summing and then input into the corresponding layer of UNet.  The Pose-Adapter can be formulated as
\begin{equation}
~\left \{ F_{pose}^{1},F_{pose}^{2},F_{pose}^{3},F_{pose}^{4}  \right \} =F_{P-AD} (I_{smplx}^{1:N},I_{skeleton}^{1:N} )
\end{equation}
where~$F_{pose}^{i}, i=1,2,3,4$ are the extracted multi-scale pose features, ~$F_{P-AD}$ is Pose-Adapter function, ~$I_{smplx}^{1:N}$ are multi-view SMPLX images and ~$I_{skeleton}^{1:N}$ are multi-view skeleton images, ~$N$ is number of views.

\textbf{ID-Adapter-G.} The face occupies a very small area in the reference image, which makes it difficult for the model to capture the information about the face, resulting in distortion of the face in the generated multi-view image. To solve this problem, We design ID adapter-G to extract facial features. Inspired by~\cite{wang2024instantid}, ID adapter-G uses a face encoder to extract face ID embedding from reference face image, which can provide fine-grained facial features. The face embedding are then injected into the UNet through a projection layer. We add global cross-attention and local cross-attention to the self-attention layer of SDXL to fuse the body features of the reference image extracted from the referencenet and facial features with multi-view features. To improve the consistency of multiple views, we added row-wise self-attention~\cite{li2024era3d} to the self-attention layer to enable effective information exchange between views. Like MV-Adapter~\cite{huang2024mv}, we adopt a parallel structure between different attention layers, which can decouple the attention layers and preserve the prior knowledge of SDXL. The attention process can be formulated as

\begin{equation}
\begin{aligned}
&h_{out}=SelfAttn(h_{in}) \\
&+GlobalCrossAttn(h_{in},h_{ref}) \\
&+LocalCrossAttn(h_{in},h_{face}) \\
&+RowWiseAttn(h_{in})+h_{in} \\
\end{aligned}
\end{equation}
where ~$h_{in}$ refers to the input hidden states in the self attention layer, ~$h_{ref}$ refers to the reference image hidden states, ~$h_{face}$ refers to the face hidden states.



\begin{figure*}[t]
\centering
\includegraphics[width=1.0\textwidth]{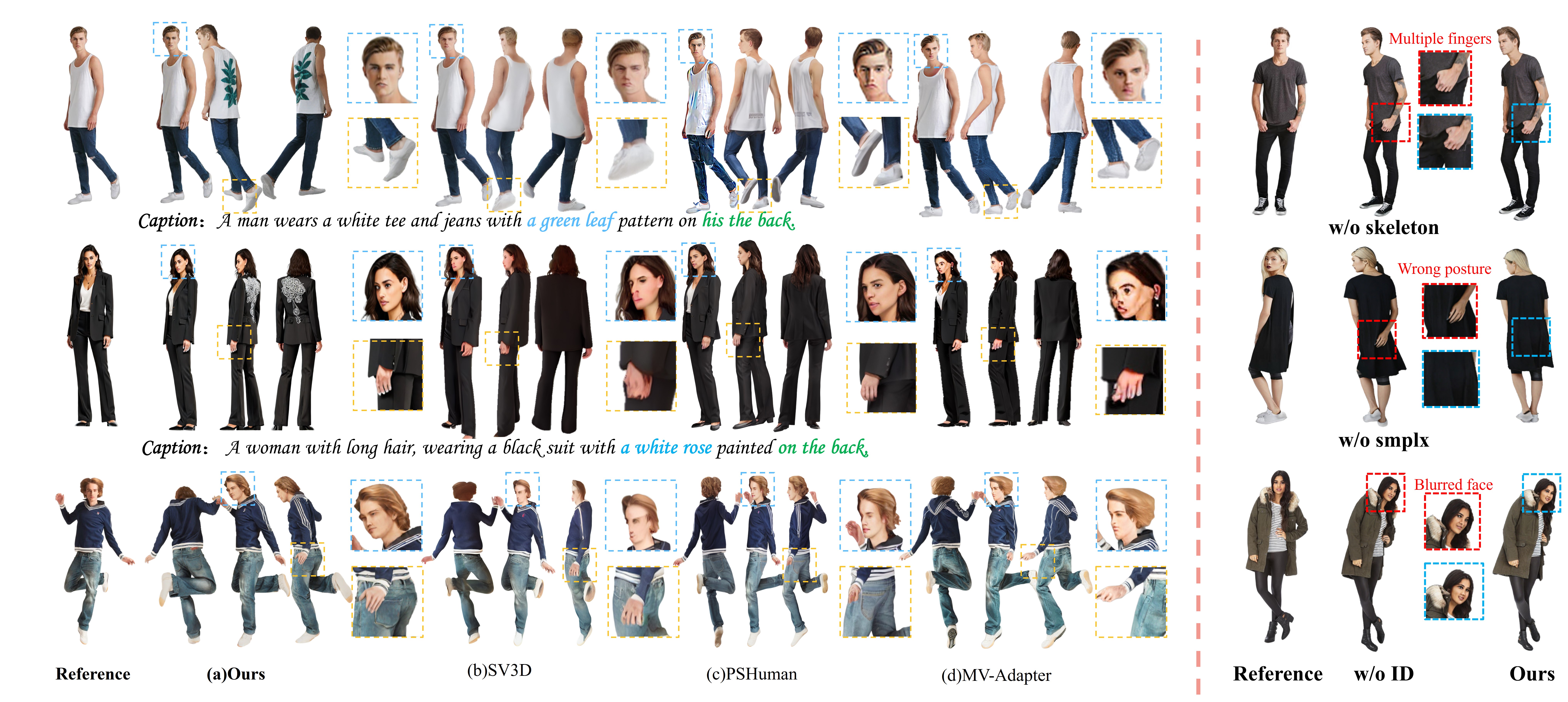} 
\caption{Qualitative comparison of comparison experiments on multi-view human image generation.}
\label{fig:multi_view_generation}
\vspace{-14pt}
\end{figure*}

\begin{figure}[t]
\centering
\includegraphics[width=1.0\columnwidth]{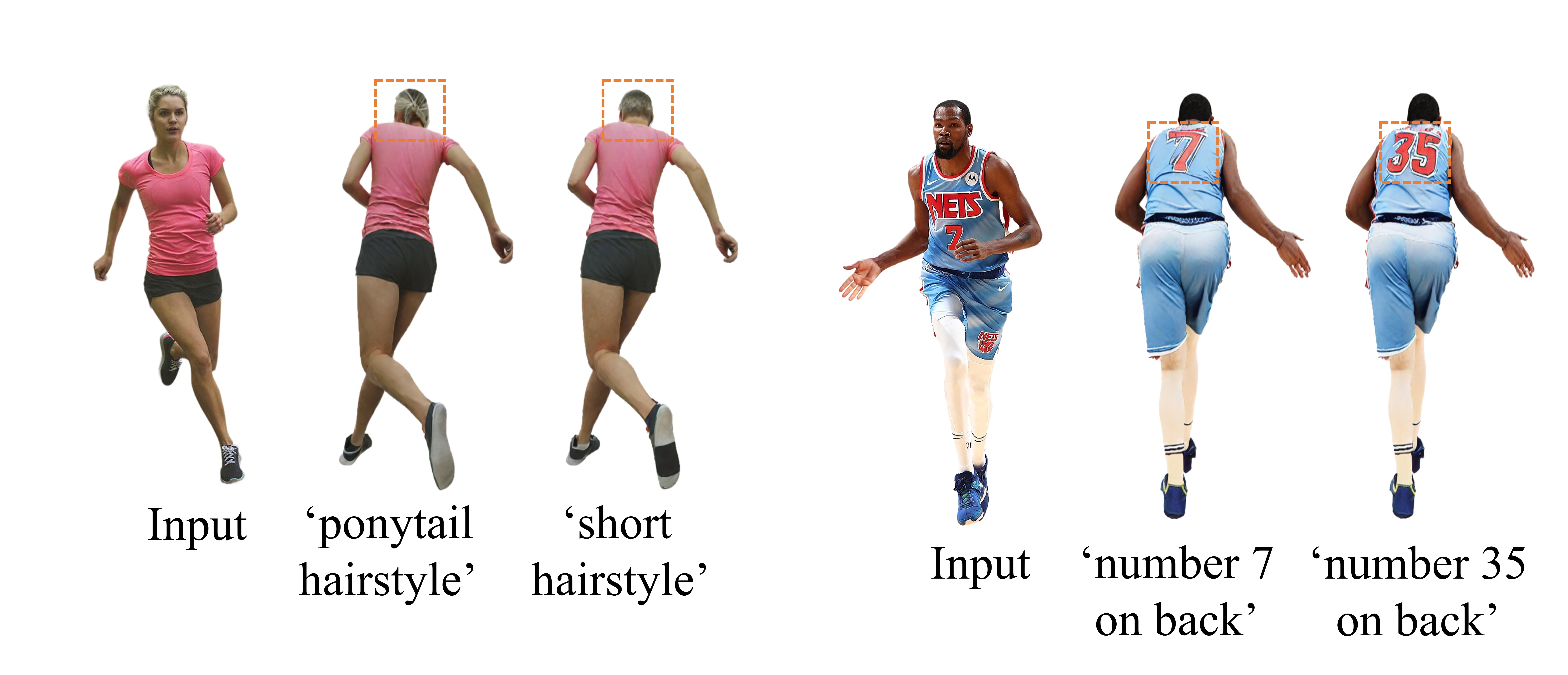}
\caption{The Text Control Results of Multi-view Generation}.
\label{fig:Tetx_Control}
\end{figure}

\begin{figure*}[t]
\centering
\includegraphics[width=1.0\textwidth]{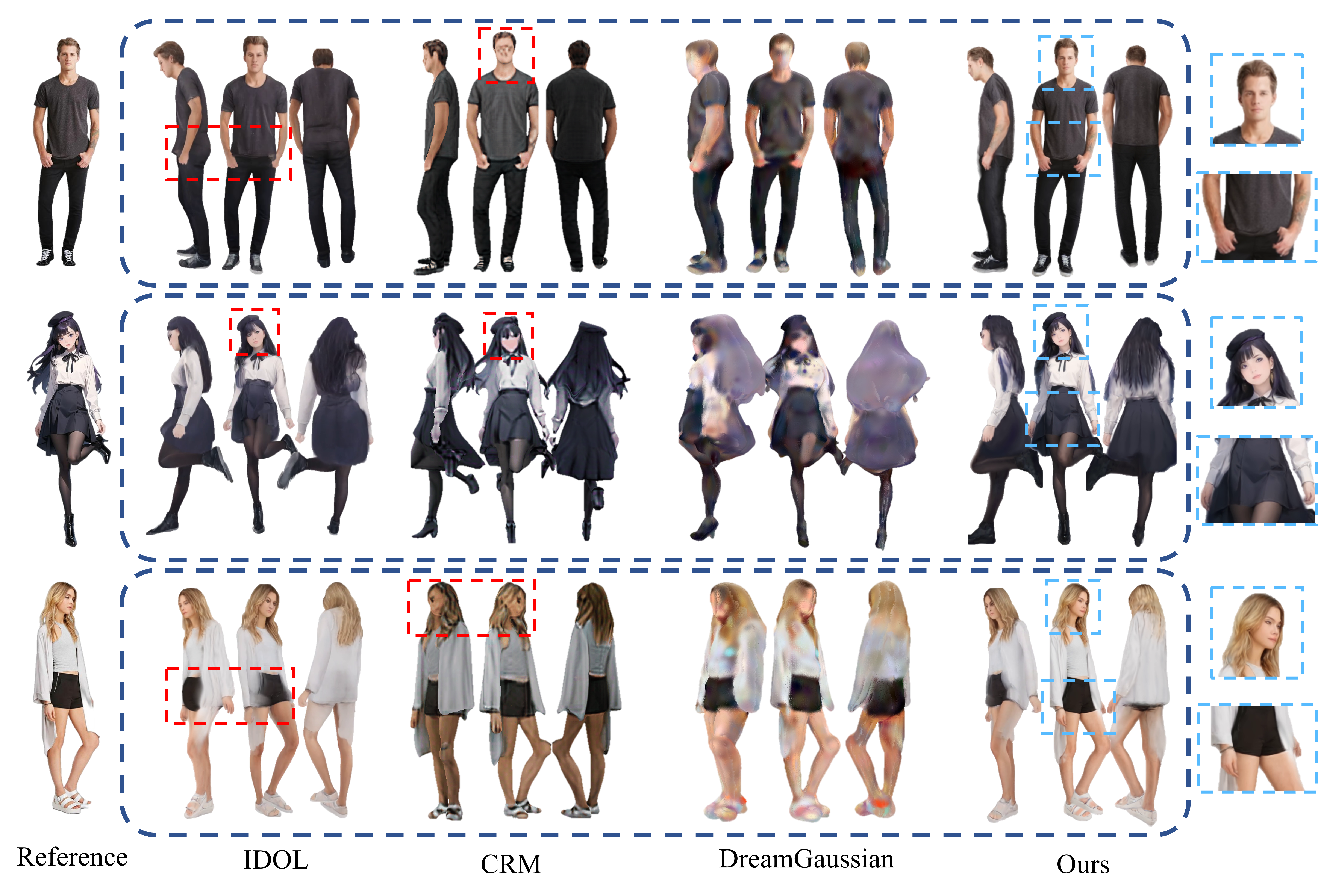} 
\caption{Qualitative comparison of 3D human reconstruction}
\label{fig:3d_human_generation}
\end{figure*}

\subsection{Multi-view Transformer for 3D Recovery}
\label{sec:Multi-view_Transformer_for_3D_Recovery}
As shown on the right side of Figure~\ref{fig:fig2}, we reconstruct high-fidelity 3D human avatars represented by 3D Gaussian Splatting from the generated multi-view images. The reconstruction process is powered by a feedforward Transformer-based model that integrates multi-view features and maps them into a 2D UV space defined by the SMPL-X model. Each 3D Gaussian primitive is associated with attributes—including color, opacity, scale, rotation, and position—determined within this UV space, which provides strong geometric priors for body modeling.

To enhance cross-view feature alignment, we design a dedicated multi-view fusion module that facilitates spatial correspondence and information aggregation across different viewpoints. The use of multi-view inputs significantly mitigates the ill-posedness introduced by monocular images by reducing occluded and invisible regions.

Furthermore, to refine the reconstruction of high-frequency facial details, we incorporate an identity-aware module, ID-Adapter-R, which injects high-resolution facial features into the Transformer pipeline. This integration substantially improves both identity preservation and fine-grained detail recovery in the final 3D output.

\subsubsection{Multi-view Body Feature Fusion}
\label{sec:Multi-view_Body_Feature_Fusion}
To address the challenges of occlusions and ambiguities in body geometry caused by monocular input, we propose a lightweight multi-view body feature fusion module~(MVBF), as shown on the left side of Figure~\ref{img:multi_view_fusion}. The module includes parallel spatial attention and view attention mechanisms, where the former focuses on intra-view relationships to enhance body part consistency, while the latter aligns features across views to resolve ambiguities and occlusions. The outputs of both attention mechanisms are fused and weighted through a learnable fusion gate, which adaptively adjusts the contribution of spatial and view features. Finally, the fused feature map is passed to a feedforward Transformer for high-quality 3D reconstruction in the UV space. By efficiently aggregating multi-view features, our module effectively mitigates the ill-posed nature of monocular image reconstruction, enabling consistent and detailed 3D human avatar generation.

\subsubsection{Identity-aware 3D Reconstruction}
\label{sec:Identity-aware_3D_Reconstruction}
Due to the small size of the facial region, the generated 3D avatar often exhibits artifacts on the face. To achieve high-fidelity identity preservation, we introduce ID-ADAPTER-R, shown on the right side of Figure~\ref{img:multi_view_fusion}, which injects high-resolution facial features into the 3D generation pipeline. First, a feature expansion network maps the facial features to the target UV space feature dimensions. Then, a gated feature fusion mechanism integrates the facial features with the UV space, enhancing facial details while preserving the original body features. Additionally, we inject facial features only in the latter half of the feedforward Transformer, as the first half mainly handles the low-frequency information related to body geometry reconstruction, while facial details belong to high-frequency information. The selective feature injection strategy ensures fine-grained identity preservation while effectively reducing computational overhead.

\subsubsection{Representation for 3D Avatars}
\label{sec:Representation_for_3D_Avatars}
Similar to previous works~\cite{zhuang2025idol,zhang20243gen,hu2024gaussianavatar}, we represent the human body using 3D Gaussian Splatting~\cite{kerbl20233d} and initialize the Gaussian primitives with SMPL-X vertices. Formally, each Gaussian primitive is defined as $\mathcal{G}_{k}=\left\{\mu_{k}, \alpha_{k}, \mathbf{r}_{k}, \mathbf{s}_{k}, \mathbf{c}_{k}\right\}$, where $\mu_{k}$, $\alpha_{k}$, $\mathbf{r}{k}$, $\mathbf{s}{k}$, and $\mathbf{c}_{k}$ denote the position, opacity, rotation, scale, and color of the Gaussian sphere, respectively. Each Gaussian primitive is mapped to a corresponding pixel in the 2D UV space, and its attributes are indirectly regressed by predicting the properties on the UV plane. This design significantly reduces computational complexity and, with the geometric prior provided by SMPL-X, enables efficient generation of 3D human animations. Given a target pose represented by SMPL-X parameters, we update the positions and rotations of all Gaussian primitives via Linear Blend Skinning (LBS) while keeping other attributes unchanged, enabling fast and flexible pose transformations of 3D avatars.

\subsection{Training Strategy}
\label{sec:Training_Strategy}
\subsubsection{Multi-view Image Generation}
In the multi-view image generation stage, we take SMPL-X images, skeleton images, reference images, reference face images and text as input samples. During training, we only train Pose-Adapter and ID-Adapter-G and freeze the parameters of the pre-trained SDXL. The objective function can be defined as:
\begin{equation}
\mathbb{E}_{z_{0},\epsilon\sim \mathcal N(0,I),c,t\  }\left \| \epsilon -\epsilon _{\theta }(z_{t},c,t )  \right \|^{2}
\end{equation}
where $c=\left \{  c_{t},c_{r},c_{f},c_{p} \right \}$,  $c_{t}$ represents text, $c_{r}$ represents reference image, $c_{f}$ represents face image, $c_{p}$ represents SMPL-X images and skeleton images, and $\theta$  represents the parameters of Pose-Adapter and ID-Adapter-G.

Our experiments show that ID-Adapter-G is overly dependent on the features of the reference image, which limits the learning of face image features. To address this issue, we randomly mask the face area of the reference image during training. Specifically, we mask the face area of the reference image with a probability of 30\% to encourage the model to make more use of face image features. To enable class free guidance during inference, we randomly drop text conditions, reference image conditions, face conditions, and pose conditions with a probability of 10\% during training.



\subsubsection{Multi-view Transformer for 3D Recovery}
In the second stage, we treat the multi-view images~$\mathbf{I}_{\mathrm{body}}^{N}$, the high-resolution facial crop from the frontal view~$\mathbf{I}_{\mathrm{face}}$, and the corresponding SMPL-X and camera parameters as a complete input sample. These inputs are jointly fed into a feedforward Transformer model to generate a canonical 3D human template. The template is then deformed according to the SMPL-X model and differentiably rendered under the given camera parameters $\pi_{t}$ to produce multi-view projections~$\hat{\mathbf{I}}_{\mathrm{body}}^N$. The loss function is computed as follows:
\begin{equation}
\begin{aligned}
\mathcal{L} = 
& \lambda_{\text{rgb}} \, \mathcal{L}_{\text{rgb}}(\mathbf{I}_{\mathrm{body}}^{N}, \hat{\mathbf{I}}_{\mathrm{body}}^N) \\
&+ \lambda_{lpips} \, \mathcal{L}_{lpips}(\mathbf{I}_{\mathrm{body}}^{N}, \hat{\mathbf{I}}_{\mathrm{body}}^N)\\
&+ \lambda_{\text{face}} \, \mathcal{L}_{lpips}(\mathbf{I}_{\mathrm{face}}, \hat{\mathbf{I}}_{\mathrm{face}})
\end{aligned}
\end{equation}
where $\lambda_{\text{rgb}}$, $\lambda_{lpips}$, and $\lambda_{\text{face}}$ are weighting coefficients that balance the contributions of each loss component.


%% file: tex/4_experiments.tex
\section{Experiments}
\subsection{Experimental Setting}
\subsubsection{Dataset}
To fine-tune Dream3DAvatar, we conducted experiments on several 3D human datasets and video data, including THuman2.1~\cite{zheng2019deephuman}, HuGe100K~\cite{zhuang2025idol}, and partial video sequences with 3D annotations from Human4DiT~\cite{shao2024human4dit}. Specifically, for 3D human data, we rendered 72 uniformly distributed views at two resolutions, $768\times768$ and $896 \times640$. Video data were cropped and resized to the same resolutions. In addition, SMPL-X parameters were estimated using Multi-HMR~\cite{baradel2024multi}. The two stages were trained with different data configurations. In the multi-view image generation stage, one reference image containing the full face was randomly selected, and six target views at corresponding intervals were used for training. In the 3D human reconstruction stage, four randomly selected images, including at least one frontal view, were used for self-supervised optimization. For quantitative evaluation, the last $50$ samples from HuGe100K, THuman2.1, and Human4DiT were used as the test set. In addition, real-world examples collected from the internet were employed for qualitative comparison.

\subsubsection{Implementation Details}
Dream3DAvatar was fine-tuned on four NVIDIA $A800$ GPUs. We adopted the AdamW optimizer~\cite{kinga2015method}, with initial learning rates set to $5\times10^{-5}$ and $1\times10^{-5}$ for the two stages, respectively. The multi-view generation and 3D reconstruction stages were trained for $40k$ and $10k$ iterations, respectively, with a mini-batch size of $1$. For evaluation, generated multi-view images were compared with ground truth using Learned Perceptual Image Patch Similarity~(LPIPS)~\cite{zhang2018unreasonable}, Peak Signal-to-Noise Ratio~(PSNR), and mean squared error~(MSE) as quantitative metrics. The total training time was approximately $4$ hours for the multi-view generation stage and $10$ hours for the 3D reconstruction stage. More details on data pre-processing and implementation are provided in the supplementary material.

\subsection{Multi-view Image Generation}
\begin{table}[t]
\centering
\resizebox{0.45\textwidth}{!}{%
\begin{tabular}{lcccc}
\toprule
\textbf{Method} & \textbf{MSE} $\downarrow$ & \textbf{PSNR} $\uparrow$ & \textbf{SSIM} $\uparrow$ & \textbf{LPIPS} $\downarrow$ \\
\midrule
SV3D & 0.0352 & 15.68 & 0.8500 & 0.2008 \\
MagicMan       & 0.0260 & 19.22 & 0.8733 & 0.1549 \\
PSHuman         & 0.0149 & 20.23 & 0.8958 & 0.1087 \\
MV-Adapter       & 0.0298 & 19.47 & 0.8688 & 0.1881 \\
\midrule
w/o face guider & 0.0080 & 21.11 & 0.9185 & 0.0863 \\
w/o skeleton guider&0.0092 & 20.47&0.9167& 0.0891 \\
w/o smplx guider & 0.0094 & 20.36 & 0.9161&0.0906\\
\midrule 
\textbf{Ours}     & \textbf{0.0052} & \textbf{22.98} & \textbf{0.9277} & \textbf{0.0711} \\
\bottomrule
\end{tabular}
}
\caption{Quantitative comparison of multi-view human image generation and ablation experiments.}
\label{tab:multi_view_generaiton}
\end{table}


For multi-view generation, we compared our method with four state-of-the-art baseline methods, including SV3D~\cite{voleti2024sv3d}, PSHuman~\cite{li2025pshuman}, Magicman~\cite{he2025magicman}, and MVAdapter~\cite{huang2024mv}. For the qualitative results, as shown in left of figure~\ref{fig:multi_view_generation}, SV3D performs poorly in terms of detail preservation and multi-view consistency. PSHuman is human-specific multi-view generation methods. It introduces body-face diffusion, which retains facial information, but still has defects in some detailed parts such as hands. Due to the lack of human body priors, MV-Adapter has deformities in human body geometry and face. Our method can generate high-quality, consistent, and identity-preserving multi-view images. As for the quantitative results, as shown in table~\ref{tab:multi_view_generaiton}, we evaluate our method and baselines on a subset of THuman2.1, and the results show that our method outperforms the existing baseline methods in all four metrics. We attribute this to our designed pose-adapter that provides body geometry information and ID-adapter-g that provides face information.

\subsection{3D avatar reconstruction}
\begin{table}[t]
\centering
\begin{tabular}{lccc}
\toprule
\textbf{Method} & \textbf{MSE} $\downarrow$ & \textbf{PSNR} $\uparrow$ & \textbf{LPIPS} $\downarrow$ \\
\midrule
DreamGaussian & 0.042 & 14.642  & 1.654 \\
SIFU              & 0.054 & 13.897 & 1.757 \\
CRM              & 0.031 & 16.211 & 1.592 \\
IDOL      & 0.011   & 20.963 & 1.289  \\
\midrule 
\textbf{Ours}                     & \textbf{0.009} & \textbf{21.322} & \textbf{1.097} \\
\bottomrule
\end{tabular}
\caption{Quantitative comparison of 3D avatar reconstruction.}
\label{tab:3d_reconstruction}
\end{table}

We evaluated our method against four representative baselines on the 3D avatar reconstruction task from both qualitative and quantitative perspectives. DreamGaussian~\cite{tang2023dreamgaussian} leverages 2D diffusion priors with score distillation sampling to iteratively refine 3D representations and adopts a progressive Gaussian densification strategy for faster convergence. CRM~\cite{wang2024crm} reconstructs meshes under multi-view supervision using RGB images and normal maps, achieving moderate improvements in multi-view consistency. SIFU~\cite{zhang2024sifu} rely on iterative geometric optimization and pixel-level alignment to enhance fine-grained details and high-fidelity reconstruction. IDOL~\cite{zhuang2025idol} employs a feedforward Transformer to directly predict 3D representations from a single image. As shown in Table~\ref{tab:3d_reconstruction}, Dream3DAvatar consistently outperforms these baselines across all metrics. Qualitative comparisons in Figure~\ref{fig:3d_human_generation} further demonstrate its superior reconstruction performance across various human types and challenging poses. It is worth noting that, for the sake of fairness, our method also uses single images for comparison instead of multi-view images. It can be seen that Dream3DAvatar achieves impressive facial detail preservation, while the baseline method struggles to recover high-frequency details and maintain texture consistency.
\subsection{Ablation Study and Analysis}

\subsubsection{Multi-view Image Generation}
The right of Figure~\ref{fig:multi_view_generation} illustrates the contribution of each module to multi-view image generation. The proposed Pose-Adapter provides strong geometric constraints, enabling pose-consistent synthesis even under challenging postures, where most baselines fail. Additional skeletal information improves the reconstruction of key body parts, such as hands. Furthermore, ID-Adapter-G significantly enhances facial details, ensuring better identity consistency across views.

\subsubsection{3D avatar reconstruction}
In the 3D reconstruction stage, we explore the effects of the number of viewpoints and ID-Adapter-R. Increasing the number of viewpoints helps cover more occluded areas, significantly improving the consistency of global geometry and texture. Removing ID-Adapter-R, on the other hand, results in a loss of facial details and degradation of identity features, validating its critical role in the fusion of high-frequency facial information. More results can be found in the appendix.

%% file: tex/5_conclusion.tex
\section{Conclusion}
In this study, we propose \textbf{Dream3DAvatar}, a novel two-stage framework for efficient, controllable, and high-fidelity 3D human avatar reconstruction from a single image. By mitigating the ill-posedness inherent in monocular reconstruction and enhancing controllability, our method effectively bridges the gap between generative diversity and reconstruction efficiency. Specifically, we design a lightweight multi-view generation module based on SDXL, incorporating geometric and semantic constraints to achieve view-consistent image synthesis. This is followed by a feedforward Transformer network equipped with an ID Adapter to further improve the accuracy and detail of 3D reconstruction. Extensive experiments demonstrate that our method achieves state-of-the-art performance on multiple benchmarks for both multi-view image synthesis and 3D reconstruction. In future work, we plan to explore driving expressive facial animations, beyond just body motion generation.
